%% file: main.tex
\pdfoutput=1

\documentclass[11pt]{article}

\usepackage[preprint]{acl}

\usepackage{times}
\usepackage{latexsym}

\usepackage[T1]{fontenc}

\usepackage[utf8]{inputenc}

\usepackage{microtype}

\usepackage{inconsolata}

\usepackage{graphicx}
\usepackage{tcolorbox}
\usepackage{enumitem}
\usepackage{amsmath}
\usepackage{float}
\usepackage{tabularx}
\usepackage{xcolor}
\definecolor{darkred}{rgb}{0.55, 0.0, 0.0}
\input{dfn}

\title{Towards More Accurate US Presidential Election via Multi-step Reasoning with Large Language Models}

\author{
  \textbf{Chenxiao Yu\textsuperscript{1}}, 
  \textbf{Zhaotian Weng\textsuperscript{1}}, 
  \textbf{Yuangang Li\textsuperscript{1}}, 
  Zheng Li\textsuperscript{2}, 
  Xiyang Hu\textsuperscript{3\textsuperscript{†}}, 
  Yue Zhao\textsuperscript{1\textsuperscript{†}} \\
  \textsuperscript{1}University of Southern California \quad
  \textsuperscript{2}Arima \quad
  \textsuperscript{3}Carnegie Mellon University \\
  \texttt{\{cyu96374, wengzhao, 
yuangang, yzhao010\}@usc.edu}, \\\texttt{winston@arimadata.com}, \texttt{xiyanghu@cmu.edu} \\
  \textsuperscript{†}Corresponding authors
}

\begin{document}
\maketitle



\begin{abstract}
\input{00abstract}

\end{abstract}


\begin{tcolorbox}[
    colback=white, 
    colframe=gray!70, title=Important Notice, 
    boxrule=0.75pt, rounded corners, 
    width=\columnwidth, before skip=10pt, after skip=10pt, 
    fonttitle=\bfseries, coltitle=white, colbacktitle=gray!80,
    left=5pt, right=5pt, top=5pt, bottom=5pt
]
\begin{itemize}[leftmargin=*,
    itemsep=2pt, 
    parsep=0pt,  
    topsep=0pt,  
    partopsep=0pt] 
    \item \textbf{Ongoing Work:} This research is ongoing as of \today.
    \item \textbf{Research Integrity:} This research is conducted independently w/o any funding.
    \item \textbf{Content Warning:} This paper may contain some offensive content generated by LLMs.
    \item \textbf{Disclaimer:} This study explores LLMs' capacity in election forecasting. Predictions do not reflect the authors' views and should not be interpreted as definitive forecasts.
\end{itemize}
\end{tcolorbox}

\input{01intro}
\input{02related}
\input{03method}

\input{05conclusion}
\bibliography{references}





\end{document}

%% file: dfn.tex
\newtcolorbox{promptbox}{
    colback=whitegreen, 
    colframe=black,      
    boxrule=0.75pt,       
    arc=3pt,             
    left=6pt,            
    right=6pt,           
    top=5pt,             
    bottom=5pt,          
    boxsep=2pt,          
    fontupper=\sffamily\small  
}

\usepackage{tikz} 

\definecolor{lightblue}{rgb}{0.68, 0.85, 0.9}
\definecolor{lightred}{rgb}{1.0, 0.8, 0.8}
\definecolor{lightgreen}{RGB}{144, 238, 144}
\definecolor{darkgreen}{RGB}{0, 100, 0}
\definecolor{whitegreen}{RGB}{248, 250, 245}

%% file: 00abstract.tex
Can Large Language Models (LLMs) accurately predict election outcomes? While LLMs have demonstrated impressive performance in various domains, including healthcare, legal analysis, and creative tasks, their ability to forecast elections remains unknown. 
Election prediction poses unique challenges, such as limited voter-level data, rapidly changing political landscapes, and the need to model complex human behavior. 
To address these challenges, we introduce a multi-step reasoning framework designed for political analysis. 
Our approach is validated on real-world data from the American National Election Studies (ANES) 2016 and 2020, as well as synthetic personas generated by the leading machine learning framework, offering scalable datasets for voter behavior modeling. 
To capture temporal dynamics, we incorporate candidates' policy positions and biographical details, ensuring that the model adapts to evolving political contexts. Drawing on Chain of Thought prompting, our multi-step reasoning pipeline systematically integrates demographic, ideological, and time-dependent factors, enhancing the model’s predictive power. 

%% file: 01intro.tex
\section{Introduction}

Large Language Models (LLMs) have demonstrated remarkable capabilities across various domains, including natural language understanding, content generation, etc. \cite{brown2020language}. 
Their potential extends far beyond mere text processing, including a broad spectrum of applications from medical diagnostics \cite{zhang2022large} to legal analysis \cite{chalkidis2022lexglue} and creative domains \cite{yang2022survey}. 
This versatility stems not only from LLMs' ability to understand and generate text but also from their capacity to leverage large amounts of common knowledge \cite{roberts2020much}, simulate diverse personas \cite{hu-collier-2024-quantifying}, and effectively model human behavior in complex social science tasks \cite{bommasani2021opportunities}. 
Specifically, LLMs have shown promising results in capturing human-like common sense reasoning \cite{zhou2020evaluating, alkhamissi2022review} and have been successfully applied to simulate human decision-making in various contexts \cite{zhou2023large, ziems2024can}. 
These multifaceted capabilities position LLMs as potential tools for simulating human decision-making processes in complex contexts. 
Recent research has begun exploring LLMs' political science applications, analyzing policy documents, campaign speeches, and public sentiment \cite{xu2022deep, haq2023large}. 
While the text-based nature of political data certainly aligns with LLMs' strengths, it is the models' holistic combination of language understanding, knowledge integration, and human-like reasoning that truly underscores their potential for simulating complicated dynamics of political-related decision-making \cite{Argyle_Busby_Fulda_Gubler_Rytting_Wingate_2023, Bisbee_Clinton_Dorff_Kenkel_Larson_2024}.

\begin{figure*}[!ht]
    \centering
    \includegraphics[width=\textwidth]{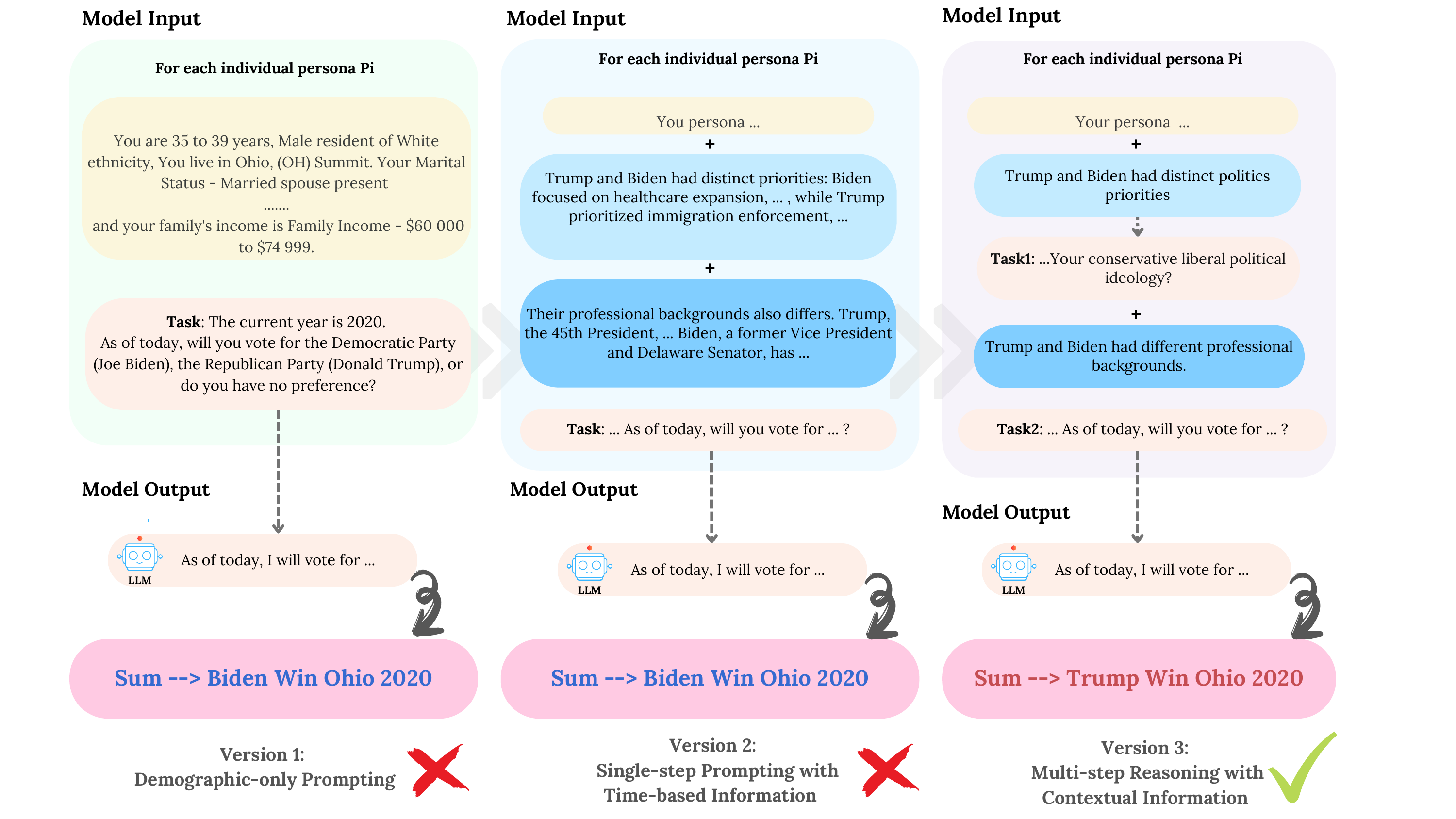}  
    \caption{Demonstration of three prompt designs in \S \ref{subsec:overview}. V1 is the direct prompt on voter demographic information, while V2 introduces time-dependent information to capture candidates' agenda and V3 also uses multi-step reasoning.
    In this example for 2020 Ohio result prediction, only V3 can accurately predict the results, demonstrating the importance of leveraging both time-dependent information and multi-step reasoning for election result prediction.}
    \vspace{-0.2in}
    \label{fig:comparision}
\end{figure*}

\noindent \textbf{Motivation}.
Despite LLMs' success in the above straightforward political science tasks, their capacity to handle more complex tasks like election prediction remains uncertain \cite{lerer2022political}.
Indeed, the potential for LLMs to accurately predict election results is an intriguing prospect, given their ability to process vast amounts of historical information and their success in other predictive tasks. However, election forecasting presents unique challenges that test the limits of LLM capabilities.
First, the high cost of acquiring voter-level data makes conducting experiments and verifying models in election prediction research challenging. 
Second, unlike many other predictive tasks, election forecasting requires modeling individual voter behavior as well as the candidates', which is inherently difficult and shifting with time. 
It remains uncertain whether text-based data alone can capture this complexity \cite{graefe2014accuracy}. 
Third, accurate election forecasting requires reasoning beyond simple inference, integrating multiple factors such as economic trends, political events, and demographic changes \cite{holbrook2016forecasting}. The capacity of LLMs to perform sophisticated reasoning for accurate election predictions is an open question \cite{wei2022chain}.

\noindent
\textbf{Our Solution}.
To address the challenges of using LLMs for election predictions, we propose a novel approach that leverages their strengths while mitigating limitations in data availability, time-varying factors, and complex political dynamics. 
First, to overcome the scarcity of detailed voter-level data, we employ the \textit{Sync} synthetic data generation framework \cite{li2020sync}, which probabilistically reconstructs individual-level demographic and behavioral profiles from aggregated public datasets. We complement this synthetic data with real-world datasets, such as the American National Election Studies (ANES) 2020 Time Series \cite{ANES2020}, ensuring our approach reflects real voting behaviors.
Second, our solution adapts to evolving political contexts by incorporating time-dependent factors. Specifically, we aggregate information from presidential campaign data, such as candidates' policy agendas and biographical backgrounds, to align our model with changing political landscapes \cite{holbrook2016forecasting}.
Third, we introduce a multi-step reasoning framework tailored for election prediction. Inspired by Chain of Thought prompting \cite{wei2022chain}, this framework decomposes the prediction process into intermediate steps, enabling the model to systematically integrate demographic information, ideological alignment, and time-sensitive factors. This multi-step reasoning improves the model's accuracy by addressing biases and overfitting issues observed with simpler approaches.
Each component of our framework builds progressively based on observations and refinements. As shown in Figs. \ref{fig:comparision} and \ref{fig:example}, we iterate through multiple pipeline versions to develop our final pipeline. This final version demonstrates significant improvements in both predictive accuracy and alignment with real-world results (Figs. \ref{fig:validation-anes}, ~\ref{fig:example-figure},  and \ref{fig:validation-state}), outperforming other pipelines across all.

Our technical contributions include:

\begin{enumerate}
    \item \textbf{First Large-Scale LLM-based Election Prediction Framework}.  
    This work establishes a new frontier in election forecasting by demonstrating how LLMs can model voter behavior using a combination of real-world data and synthetic datasets, capturing voter-level dynamics with significant scale and detail.

    \item \textbf{Novel Multi-Step Reasoning Framework}.  
    We introduce a novel multi-step reasoning process tailored specifically for political forecasting. 
    This framework enhances the model’s ability to integrate and analyze over 11 critical, time-sensitive features—such as policy agendas and candidates' backgrounds—allowing for more complex, context-aware predictions.

    \item \textbf{New Insights and Future Directions}.  
    Our analysis uncovers essential insights into the strengths and limitations of LLMs in election prediction, including potential ideological biases and the challenges of temporal modeling. 
    Future research will explore the integration of multiple LLMs for comparative analysis and further refinement of prompting to improve prediction reliability and robustness.
\end{enumerate}


%% file: 02related.tex
\section{Related Work}

\subsection{LLMs in Political Science: A New and Emerging Field}

The application of LLMs in political science represents a new and rapidly evolving field, with a limited but growing body of research. While LLMs have revolutionized natural language processing and various other domains, their potential in political science remains largely untapped. Initial studies have demonstrated promising results in areas such as election forecasting, policy analysis, and public opinion simulation \cite{smith2023llms, johnson2024political}. However, political science often involves complex social dynamics and multi-layered causal relationships, posing significant challenges for effectively utilizing LLMs in this context \cite{brown2023challenges}.

Recent work by \citet{chen2024decoding} highlights the potential of LLMs in decoding political speeches and policy documents, emphasizing the need for domain-specific fine-tuning to capture the subtleties of political language. Future research is likely to focus on designing models that can handle the intricacies of political discourse while ensuring robustness against biases and misleading inferences \cite{wilson2024robust, thompson2023ethical}.

The latest work by \citet{li2024politicalllmlargelanguagemodels} presents the first systematic framework for integrating LLMs into computational political science. 
The authors propose a novel taxonomy that categorizes existing work into two main perspectives: political science applications and computational methodologies. From the political science perspective, they highlight LLMs' capabilities in automating predictive and generative tasks, simulating behavior dynamics, and enhancing causal inference. From the computational perspective, they detail advances in data preparation, fine-tuning approaches, and evaluation methods specifically tailored for political contexts. The paper identifies critical challenges, including the need for domain-specific datasets, addressing bias and fairness issues, incorporating human expertise effectively, and developing evaluation criteria that are aligned with political science requirements.
Our work differs from Political-LLM by giving more in-depth analysis on the Decision-Making
simulation in Political Science.

Although LLMs have seen rapid advancements, their application in political science remains limited. Only a small number of studies have explored how LLMs can be used for tasks like election prediction, policy analysis, and public opinion tracking. Political text analysis is an important area, and some early benchmark datasets are starting to emerge. However, political language is often complex, with nuanced meanings and context, which presents a significant challenge for LLMs \cite{anderson2023nuances, williams2024challenges}.

One notable example is the ``Political Campus'' project by \citet{roberts2023political}, which developed a benchmark dataset specifically for election prediction and evaluated LLM performance on various election-related questions. This work has been instrumental in highlighting both the potential and limitations of LLMs in political forecasting. Similarly, the research by \citet{kim2024sentiment} on using LLMs for policy sentiment analysis demonstrates the models' capacity to process large volumes of public opinion data, while also underscoring the need for careful interpretation of results.

%% file: 03method.tex


\section{Using LLMs for Election Result Prediction}

How can we effectively leverage LLMs to predict election results? 
In this work, we simulate \textit{each voter's decision-making process} by providing LLMs with detailed voter information and asking them to predict voter preferences based on that data. 
To achieve this, we focus on two key aspects: (1) establishing an evaluation framework with appropriate datasets that contain voter-level information, and (2) designing an LLM-based pipeline for accurate election predictions.
In \S \ref{subsec:data_verification}, we introduce the datasets used in this study and describe the details of our pipeline evaluation process. 
We then provide an overview of our design approach in \S \ref{subsec:overview}, with a discussion of three progressive pipelines in \S \ref{subsec:v1}, \S \ref{subsec:v2}, and \S \ref{subsec:v3}, ranging from simple prompting to multi-step reasoning based on observations.
Finally, we evaluate the performance of these three pipelines on two datasets in \S \ref{subsec:evaluation_results}.



\subsection{Datasets, Evaluation, and Settings}
\label{subsec:data_verification}

Before presenting the pipelines for election prediction, we first describe the datasets, establish the evaluation framework, and introduce the experimental settings. In this work, we use two data sources:  
(1) real-world American National Election Studies (ANES) 2016 and 2020 Time Series data \cite{ANES2016, ANES2020}, and  
(2) voter-level synthetic data generated using advanced machine learning techniques based on aggregated information \cite{li2020sync}.  
Both datasets provide non-personally identifiable voter-level information. 
The following sections offer detailed descriptions of these datasets, explain their role in the evaluation framework, and outline the experimental settings used for testing the pipelines.

\subsubsection{Real-world Data by American National Election Studies (ANES)}

For evaluation, we use pre-election data from the ANES 2016 and 2020 Time Series Studies \cite{DVN/JPV20K_2022, ANES2020}, which provide 4,270 and 8,280 real-world samples, respectively, from individuals who participated in the 2016 and 2020 elections. 
The dataset includes a wide range of variables: (1) racial/ethnic self-identification, (2) gender, (3) age, (4) ideological self-placement on a conservative-liberal scale, (5) party identification, (6) political interest, (7) church attendance, (8) frequency of discussing politics with family and friends, (9) patriotic feelings associated with the American flag (unavailable in 2020), and (10) state of residence (unavailable in 2020). Additionally, the dataset records how individuals voted in both the 2016 and 2020 elections.
Previous studies, such as \citet{Argyle_Busby_Fulda_Gubler_Rytting_Wingate_2023}, have evaluated GPT-3 using this dataset. We apply our method directly to this established benchmark to assess its effectiveness and performance.





\subsubsection{Synthetic Data for the US Population}

In addition to the medium-sized benchmark dataset, we utilize synthetic demographic data derived from a 1:1 synthetic population dataset of the United States \cite{li2020sync}. 
Synthetic data plays a crucial role in social and applied sciences, with recent applications in water quality estimation \cite{chia2023artificial}, financial modeling \cite{potluru2023synthetic}, tourist profiling \cite{merinov2023behaviour}, and measuring the social impact of engineered products \cite{stevenson2023creating}. 
High-quality synthetic datasets provide researchers with large-scale data at a lower cost while maintaining privacy, making them a reliable resource.

For our purposes, the synthetic data enables the creation of a cost-effective, large-scale virtual panel of respondents that is both ``wide" (each respondent has over 50k modeled features) and ``long" (enough samples to reflect a national dataset). However, running LLM inference on the entire U.S. population would be prohibitively expensive, so we employ a sampling strategy. Given the pivotal role of swing states in determining election outcomes, we focus on simulating voter behavior in these states while including representative samples from red and blue states for comparison.



\noindent \textbf{Synthetic Data Generation:}  
The synthetic data used here is generated using the \texttt{SynC} framework \cite{li2020sync}, which reconstructs individual-level data from aggregated sources where collecting real-world individual data is impractical due to privacy, time, or financial constraints. 
\texttt{SynC} is widely recognized and applied across multiple fields to support research and overcome data limitations. For instance, it has been used in outlier detection \cite{Li2020COPOD}, finance \cite{Potluru2023SyntheticFinance}, tabular data modeling \cite{Borisov2022DeepTabularSurvey}, healthcare \cite{Sichani2024SyntheticHealth}, and tourism \cite{Merinov2023Tourism}, demonstrating its effectiveness and importance in various domains.

\texttt{SynC} leverages publicly available data, such as the 2023 American Community Survey (ACS), which provides data on 242,338 census block groups, including population statistics and response proportions for each block. 
Using \textit{Data Downscaling}, \texttt{SynC} probabilistically recreates the 340 million residents represented in the aggregated census data.
For our simulation, the synthetic population includes variables relevant to election predictions: (1) age, (2) gender, (3) ethnicity, (4) marital status, (5) household size, (6) presence of children, (7) education level, (8) occupation, (9) individual income, (10) family income, and (11) place of residence.

\texttt{SynC} addresses the challenge of reconstructing individual data $\{x_{m,1}^d, \ldots, x_{m,n_m}^d\}$ from aggregated observations $X_m^d = \sum_{k=1}^{n_m} x_{m,k}^d / n_m$, where $X^d$ is the $d$-th survey question of interest, $m$ is the census block id and $n$ is the number of individuals in $m$. A \textit{Gaussian copula} is employed to model dependencies between survey questions. Given a $d \times d$ covariance matrix $\Sigma$ of the $d$ sruvey questions, the synthetic individuals are drawn as:
\begin{small}
    \begin{equation}
        Z_m^d \sim N(0, \Sigma), \quad u_m^d = \Phi(Z_m^d), \quad X_m^{d} = F^{-1}_d(u_m^d),
    \end{equation}
\end{small}
where $Z_m^d \sim N(0, \Sigma)$ denotes a random seed from a multivariate normal distribution, $\Phi$ is the cumulative distribution function (CDF) of the standard normal distribution, and $F^{-1}_d$ is the inverse CDF of the marginal distribution for feature $d$, which is estimated based on census block level data.
To maintain alignment with aggregated data, SynC uses \textit{marginal scaling}. For categorical variables, it applies a multinomial distribution:
\begin{equation}
    X^d \sim \text{Multi}(1, c^d, p_{m,k}^d),
\end{equation}
where $p_{m,k}^d$ is the probability distribution over $c^d$ categories for question $d$ and individual $k$. Marginal constraints are adjusted iteratively if discrepancies arise between sampled and target proportions.

The multi-phase \texttt{SynC} framework ensures that: (1) marginal distributions of individual features align with real-world expectations, (2) feature correlations are consistent with aggregated data, and (3) aggregated results match the input data. For further details on \texttt{SynC}’s methodology and algorithms, please see the original paper \cite{li2020sync}.

\begin{figure*}[!ht]
    \centering
    \includegraphics[width=\textwidth]{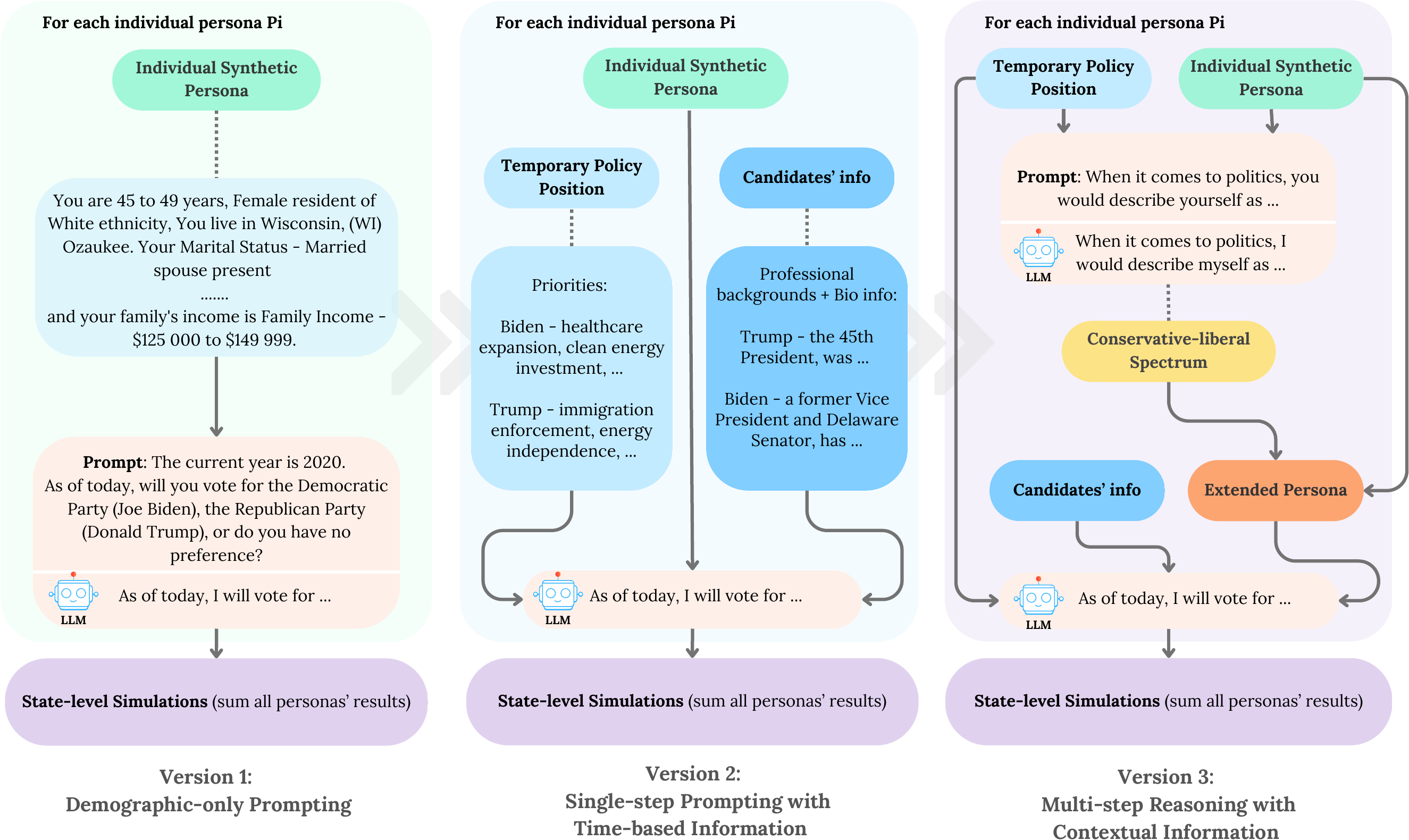}  
\caption{
Progressive design of LLM pipelines for election predictions. 
\textbf{V1: Direct Prompt on Demographic} (\S \ref{subsec:v1}) uses static demographic personas but lacks temporal context. 
\textbf{V2: Time-dependent Prompts} (\S \ref{subsec:v2}) incorporates election-year policy shifts and candidate information, but struggles with overloaded prompts that limit prediction accuracy. 
\textbf{V3: Multi-step Reasoning} (\S \ref{subsec:v3}) structures the decision-making process into sequential steps, allowing for more nuanced reasoning and yielding unbiased results that align closely with real-world outcomes. 
Each version aggregates individual results through state-level simulations to reflect broader election trends.
}
\vspace{-0.2in}
    \label{fig:example}
\end{figure*}

\noindent \textbf{Partition Design and State Categorization:}  
The synthetic dataset evaluation will operate at the state level, where we sample synthetic individuals from each state to simulate voter behavior and aggregate their votes to compare the simulated outcomes with actual election results. 
Given the critical role of swing states and tipping-point states in determining election outcomes, our primary focus is on these states, which include Florida (FL), Wisconsin (WI), Michigan (MI), Nevada (NV), North Carolina (NC), Pennsylvania (PA), Georgia (GA), Texas (TX), Minnesota (MN), Arizona (AZ), and New Hampshire (NH). 
For broader comparison in the following evaluations, we also sample from several reliably ``red states,'' such as Alabama (AL), Arkansas (AR), Idaho (ID), Ohio (OH), and South Carolina (SC), as well as from ``blue states,'' such as California (CA), Illinois (IL), New York (NY), New Jersey (NJ), and Washington (WA). 
These classifications are based on the 2020 election results as described by Wikipedia \cite{wikipedia_swing_state}.


\noindent \textbf{Sampling Method:}  
Running LLM inference on the entire synthetic population is computationally prohibitive, so we adopt a random sampling approach. Each state serves as a sampling unit, with sample sizes ranging between 1/100 and 1/2000 of the synthetic population, depending on the state’s population size. For example, a 1/2000 sampling ratio is applied to highly populated states like California, while a 1/100 ratio is used for smaller states such as New Hampshire. This approach ensures a minimum sample size of $4269$ individuals per state, corresponding to a 1.5\% margin of error at a 95\% confidence level, to maintain sufficient representation. Although our primary focus is on swing states due to their critical influence on election outcomes, we apply the same sampling method to red and blue states included in our simulations to ensure consistency across the analysis.

\subsubsection{Evaluation Using Real-world and Synthetic Datasets 
}

We employ two evaluation methods to assess our proposed approaches. 
First, for the ANES 2016 and 2020 benchmarks \cite{ANES2016, ANES2020}, we follow the methodology of \citet{Argyle_Busby_Fulda_Gubler_Rytting_Wingate_2023}. We compare the average voting probabilities:  

\begin{small}
    \begin{equation}
    \text{Probability} = \frac{\text{Republican Votes}}{\text{Republican Votes} + \text{Democratic Votes}}
    \label{eq:prob}
    \end{equation}
\end{small}

calculated across the entire sample. Accuracy is assessed by comparing the predicted winning party with the actual election outcome. 

Second, for the synthetic dataset, we treat each state as an independent validation unit. We compare the predicted results—both in terms of the winning candidate and vote share percentages—against the actual 2020 election results for each state. Accuracy is evaluated based on:  
(1) the agreement between the predicted and actual winning candidate for each state, and
(2) the aggregate performance across all states, ensuring that the model reflects overall election trends.  
This state-level evaluation uses voter-level information processed through LLMs to predict outcomes accurately.

\subsubsection{Hardware and LLM Settings}

Our experiments are deployed on a GPU server equipped with an AMD EPYC Milan 7763 processor, 1 TB (64x16 GB) DDR4 memory, 15 TB SSD storage, and 6 NVIDIA RTX A6000 Ada GPUs. 
For the LLM component, we primarily utilize OpenAI's GPT-4o model for election predictions. 
Additionally, Meta's LLaMA 3.1 405B model is used in intermediate steps to provide neutral summarization of time-dependent information, enhancing certain pipelines by capturing temporal dynamics more effectively.

\subsection{Our Progressive Design of LLM Pipelines}
\label{subsec:overview}

In this section, we present our progressive design for making voter-level election predictions using LLMs. As illustrated in Fig.~\ref{fig:example}, the methodology evolves through three distinct versions, each addressing limitations of the previous approach and incorporating more advanced techniques.

\begin{enumerate}[label=\textbf{V\arabic*:}, leftmargin=*]
    \item \textbf{Demographic-only Prompting (\S \ref{subsec:v1}):}  
    This version uses static demographic personas to prompt the LLM for voter-level predictions. While straightforward, it cannot account for temporal shifts in candidates' political focus over time, limiting its predictive power.
    
    \item \textbf{Single-step Prompting with Time-based Information (\S \ref{subsec:v2}):}  
    To address temporal factors, this version enriches the prompts with election-year-specific information, such as candidates’ policy positions and campaign focuses. However, packing all relevant info. into a single prompt creates cognitive overload, which can hinder effective reasoning and reduce prediction accuracy.

    \item \textbf{Multi-step Reasoning with Contextual Information (\S \ref{subsec:v3}):}  
    This version breaks down the prediction process into sequential steps to improve reasoning. Structuring the decision-making process allows the model to effectively incorporate voter information, candidates' profiles, and political context. Our experiments across all datasets show that this approach produces improved predictions closely aligned with real-world outcomes.
\end{enumerate}

The subsequent sections provide detailed descriptions of each version and its development. The quantitative evaluation of the three pipelines is presented in \S \ref{subsec:evaluation_results}.






\subsubsection{Version 1: Demographic-only Prompting}
\label{subsec:v1}




Building on prior research demonstrating LLMs’ ability to simulate human behavior \cite{xie2024largelanguagemodelagents}, this initial version directly prompts the LLM with a persona and asks how that persona would vote \cite{Argyle_Busby_Fulda_Gubler_Rytting_Wingate_2023}. This method provides all relevant information simultaneously, making it the simplest approach for voter-level prediction. 

\begin{promptbox}
\noindent \textbf{Task:} You are persona [age, gender, ethnicity, marital status, household size, presence of children, education level, occupation, individual income, family income, and place of residence.] The current year is [year].\\

\noindent Please answer the following question as if you were the resident:

\begin{enumerate}[leftmargin=*]
    \item As of today, will you vote for the Democratic Party (Joe Biden), the Republican Party (Donald Trump), or do you have no preference?
    \begin{itemize}[leftmargin=*]
        \item Democratic
        \item Republican
        \item No Preference
    \end{itemize}
\end{enumerate}
\end{promptbox}

In our prompt, we explicitly specify the year as 2020 to avoid confusion, since the LLM used—GPT-4o—has knowledge only up to 2023. 
Without this clarification, the model might assume the present year is 2023, impacting its predictions. The structure of the voting options follows the style used in Pew Research Center's 2014 Political Polarization and Typology Survey \cite{pew2014polarization}.


\noindent \textbf{Limitations:}  
While simple and intuitive, this approach is limited by its inability to account for temporal changes in candidates’ political agendas and public opinion. As a result, predictions for different years (e.g., 2020 vs. 2024) may not reflect meaningful variation, reducing the method’s effectiveness in dynamic election contexts.

\subsubsection{Version 2: Single-step Prompting with Time-based Information}
\label{subsec:v2}

Capturing macro-level and time-dependent variables is essential for bottom-up agent-based modeling in election prediction \cite{10.1371/journal.pone.0270194}. 
To enhance the contextual relevance of our simulations, we extended our pipeline by integrating election-year data sourced from Ballotpedia, a well-regarded political information platform. 
It includes campaign agendas, key policy stances, and candidates' biographical and professional backgrounds.

Delivering this time-based information neutrally to LLMs is crucial to avoid skewed predictions. 
Given the documented political bias in LLMs \cite{feng2023pretrainingdatalanguagemodels}, we experimented with both GPT-4o and LLaMA3-405B to summarize the information neutrally. Our preliminary findings indicate that LLaMA3-405B offers more balanced expressions 
These unbiased summaries were integrated into the prompts as follows:

\begin{promptbox}
\textbf{Task:}  
You are persona [demographics]. The current year is [year]. [Two parties' policy agenda]. [Presidential candidates' biographical and professional backgrounds].\\

Please answer the following question as if you were the resident:
\begin{enumerate}[leftmargin=*]
    \item As of today, will you vote for the Democratic Party (Joe Biden), the Republican Party (Donald Trump), or do you have no preference?
    \begin{itemize}[leftmargin=*]
        \item Democratic
        \item Republican
        \item No Preference
    \end{itemize}
\end{enumerate}
\end{promptbox}

\noindent \textbf{Limitations:}  
While this version creates more dynamic and contextually grounded simulations, it introduces a skew in the predictions for certain states. 
Specifically, when tested across five ``deep red'' states, five ``deep blue'' states, and all 11 swing and tipping-point states, we observed a pronounced skewness towards the Democratic Party, even in historically red states such as Alabama and South Carolina, as well as swing states like Texas and Florida. This aligns with prior research suggesting that GPT-4o tends towards liberal ideologies \cite{feng2023pretrainingdatalanguagemodels}.
However, this skewness was less pronounced when tested on the ANES 2020 dataset \cite{ANES2020}. In that case, the predicted share for Trump was 46.7\%, slightly higher than the ground truth of 41.2\%. 
Further analysis revealed that the ANES dataset contains more features than standard demographic datasets, including ideological self-placement along the conservative-liberal spectrum. 

Through additional experimentation, we found that removing the ideological self-placement feature from the ANES data caused the predictions to shift significantly toward demographic factors.
This suggests that ideological self-placement is critical in mitigating political skew, showing its importance as a corrective feature in election prediction.

\subsubsection{Version 3: Multi-step Reasoning with Contextual Information}
\label{subsec:v3}



To overcome the limitations of Version 2 and leverage insights from the ANES dataset analysis, we developed a multi-step prompting pipeline. Inspired by the Chain of Thought prompting strategy \cite{wei2022chain}, this approach divides the task into intermediate steps, allowing the model to process information more systematically and accurately.

The process consists of two main steps: (\textbf{1}) Conservative-Liberal Spectrum Placement: First, the LLM is provided with a specific persona along with the current policy positions of both parties. The model is then asked to place the persona on the conservative-liberal spectrum based on the provided information. 
(\textbf{2}). Extended Persona and Voting Simulation:
The conservative-liberal spectrum placement is incorporated into the persona to create an extended persona. This extended persona, along with the time-based information, is used in the second step to simulate voting behavior. The overall prompts are structured as follows:

\begin{promptbox}
\textbf{Step 1:}  
You are a persona with [demographics]. The current year is [year]. [Two parties' policy agenda].  

When it comes to politics, would you describe yourself as:

\begin{center}
\begin{tabularx}{\linewidth}{X X}
    No answer & Very liberal\\
    Somewhat liberal & Closer to liberal\\
    Moderate& Closer to conservative\\ 
    Somewhat conservative & Very conservative\\
\end{tabularx}
\end{center}

\textbf{Step 2:}  
You are a persona with [demographics]. Your [conservative-liberal spectrum]. The current year is [year]. [Two parties' policy agenda]. [Presidential candidates' biographical and professional backgrounds].  

Please answer the following question as if you were the resident:

\begin{enumerate}[leftmargin=*]
    \item As of today, will you vote for the Democratic Party (Joe Biden), the Republican Party (Donald Trump), or do you have no preference?
    \begin{itemize}[leftmargin=*]
        \item Democratic
        \item Republican
        \item No Preference
    \end{itemize}
\end{enumerate}
\end{promptbox}

We applied this multi-step method to simulate the 2020 election at the state level, covering various types of states (e.g., ``deep red'', ``deep blue'', swing, and tipping-point states). 
This approach resulted in significantly improved performance, with outcomes closely aligning with real-world scenarios. 
Specifically, the voting distributions reflected expected patterns in deep red and deep blue states such as Alabama and California, with vote counts closely matching ground-truth distributions. For swing states, the results were also accurate, except for Arizona and North Carolina, where the predicted outcomes flipped. Nonetheless, in all other swing states, the simulated vote distributions closely mirrored the real-world outcomes, capturing these states' balanced and marginal dynamics.

Based on this approach's improved accuracy and alignment with real-world results, we choose Version 3 as our \textbf{final pipeline} for election prediction. The structured, multi-step reasoning process not only mitigates skewness but also effectively captures the complex dynamics of voter behavior across different states.

\begin{figure}[!t]
    \centering
    \includegraphics[width=\linewidth]{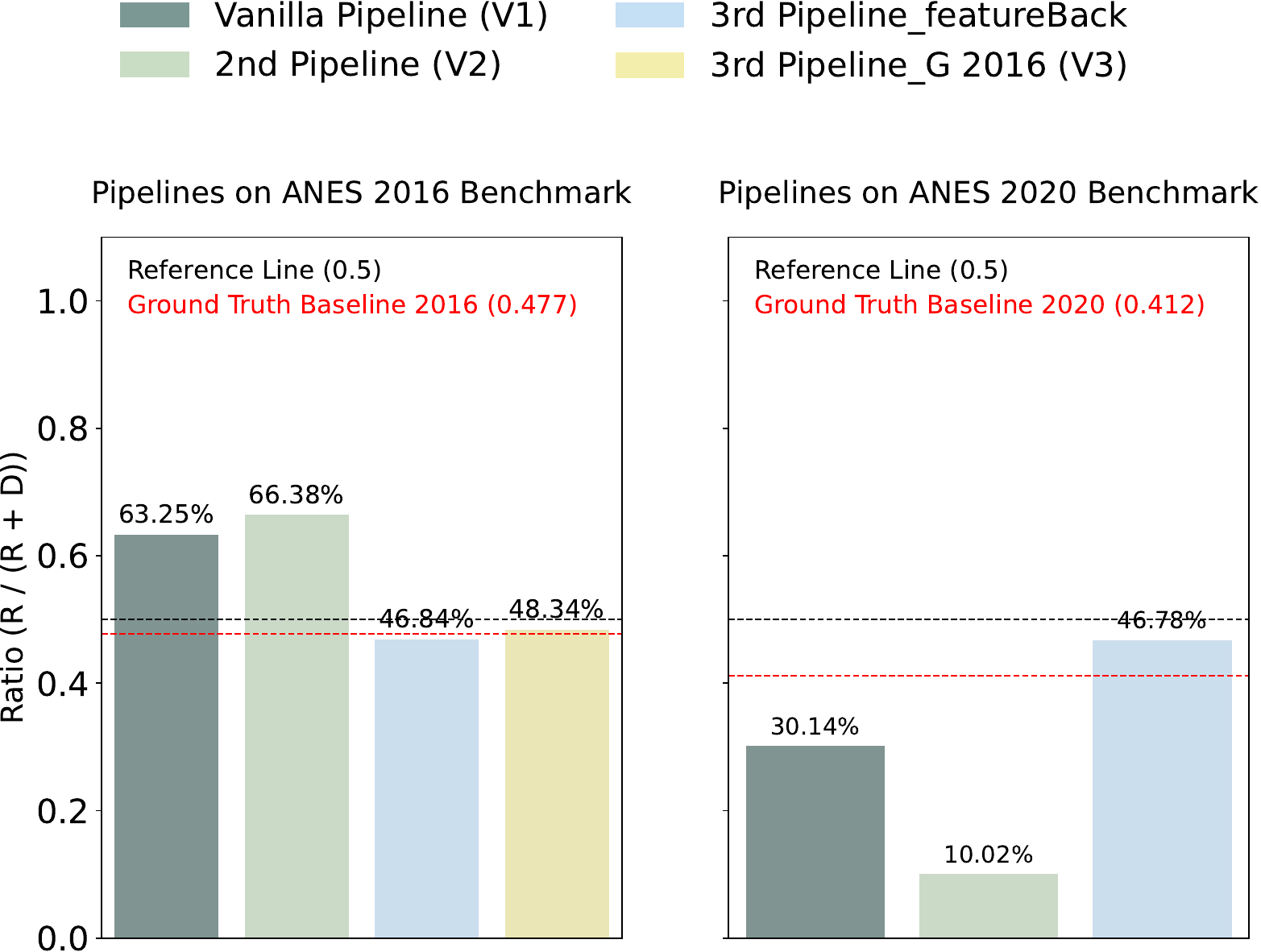}  
    \caption{Comparison of the three pipelines on ANES 2016 and 2020 benchmarks. 
    The y-axis shows the predicted Republican vote ratio (R / (R + D)), with 0.5 indicating a balanced outcome. 
    V1 (Vanilla Pipeline) and V2 (Single-step Time-based Prompting) overestimate Republican support, particularly in 2016. 
    V3 (Multi-step Reasoning) achieves the most accurate results, closely matching the ground truth ratios: 48.34\% vs. 47.7\% (2016) and 46.78\% vs. 41.2\% (2020). 
    These results highlight the improved accuracy of V3.}
    \label{fig:validation-anes}  
    \vspace{-0.2in}
\end{figure}


\subsection{Validation on the Proposed Pipelines}
\label{subsec:evaluation_results}

\subsubsection{Evaluations on Real-world Data (ANES)}

\textbf{Settings}. 
We evaluate our pipelines using the public ANES 2016 and 2020 Time Series datasets \cite{ANES2016} \cite{ANES2020} to: (1) assess the overall performance of each pipeline and (2) validate the V3 pipeline’s ability to generate the critical Conservative-Liberal Spectrum feature.

For the first two pipelines (V1: Demographic-only Prompting and V2: Single-step Prompting with Time-based Information), we manually excluded the Conservative-Liberal Spectrum feature during evaluation to simulate how these pipelines would perform based purely on demographic data. This step mimics the limitations of simpler prompts that lack deeper ideological alignment.


The V3 pipeline, as outlined in \S \ref{subsec:v3}, addresses the limitations of earlier versions by using a multi-step reasoning approach inspired by Chain of Thought prompting \cite{wei2022chain}. 
This design involves two primary steps. First, the LLM places a persona on the Conservative-Liberal Spectrum based on the persona’s demographics and the two parties' policy positions. Second, this placement is incorporated into an extended persona, which, along with policy agendas and candidates' biographical information, is used to simulate voting behavior.
The evaluation of the V3 pipeline involved slightly different strategies for the two datasets. 
The 2020 ANES dataset lacks key demographic features such as `state of residence` and `patriotic feelings associated with the American flag,’ making it more challenging for the LLM to generate the Conservative-Liberal Spectrum. To address this, we restored the original spectrum in the 2020 dataset for evaluation.
For the 2016 dataset, which contains all relevant features, we tested two variations of the V3 pipeline. In the first test, labeled \textit{3rd Pipeline\_G 2016}, the LLM generated the Conservative-Liberal Spectrum using the available demographic information. In the second test, labeled \textit{3rd Pipeline\_featureBack}, we restored the original spectrum to simulate a comparison. The results for these evaluations are shown in Figure~\ref{fig:validation-anes}.
    \vspace{-0.2in}

\begin{figure}[!ht]
    \centering
    \includegraphics[width=\linewidth]{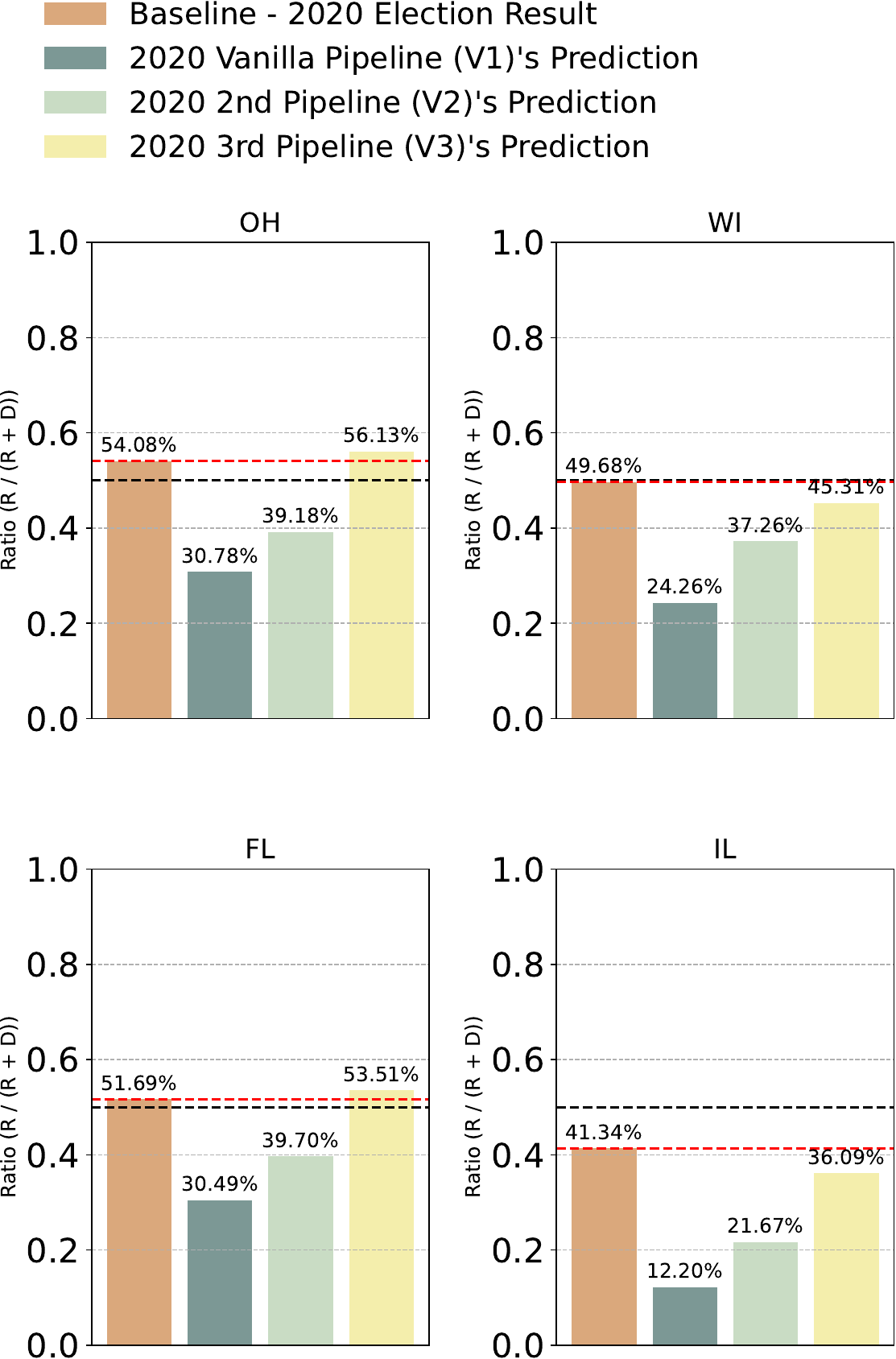}  
    \caption{LLM’s predictions for four states in the 2020 election compared with Ground Truth results. The figure presents results for one red state (Ohio, OH), one blue state (Illinois, IL), one swing state (Wisconsin, WI), and one tipping-point state (Florida, FL). 
    V1 and V2 pipelines tend to underestimate Republican support, while V3 (Multi-step Reasoning) provides the closest alignment with actual outcomes, especially in swing and tipping-point states.
    }
    \label{fig:example-figure}  
    \vspace{-1em}  
\end{figure}



\noindent
\textbf{Validation Results}.
Our results reveal a consistent trend: when the Conservative-Liberal Spectrum is removed, the predictions shift toward the winning party of the respective election year. 
For instance, in 2016, the LLM predicted 63.25\% support for Trump, favoring the Republican Party. In 2020, the prediction shifted to 30.14\% support for Trump, favoring the Democratic Party. This shift becomes even more pronounced in the V2 pipeline, where additional time-based information introduces further skew in the predictions.


Despite the missing features in the 2020 dataset, the V3 pipeline still demonstrated noticeable improvements over the simpler V1 and V2 pipelines. On the more feature-complete 2016 dataset, the V3 pipeline performed well, achieving 46.84\% support for Trump when using the original feature and 48.38\% with the generated feature—both closely aligning with the ground truth baseline of 47.7\%. 
Notably, the generated spectrum in the \textit{3rd Pipeline\_G 2016} version produced results even closer to the ground truth than the restored spectrum.
The performance improvements observed underscore the importance of incorporating ideological alignment in voter simulations.



\subsubsection{Evaluations on Synthetic Data for the 2020 US Population}  

In addition to the nationwide evaluation on the ANES datasets, we conducted state-level simulations using synthetic data to compare predictions with actual 2020 election outcomes. 
For each state, we performed random sampling based on population size to ensure a statistically meaningful number of personas. The simulation outcomes were then benchmarked against official 2020 Presidential General Election Results from the Federal Election Commission (FEC). As in the benchmark evaluations, we calculated the average voting probabilities to assess the alignment of predictions with real-world outcomes.
We evaluated five red states, five blue states, and 11 swing and tipping-point states. Figure~\ref{fig:example-figure} highlights representative results from these categories, providing insights into the model’s performance in different electoral contexts.

\begin{figure}[!t]
    \centering
    \includegraphics[width=\linewidth]{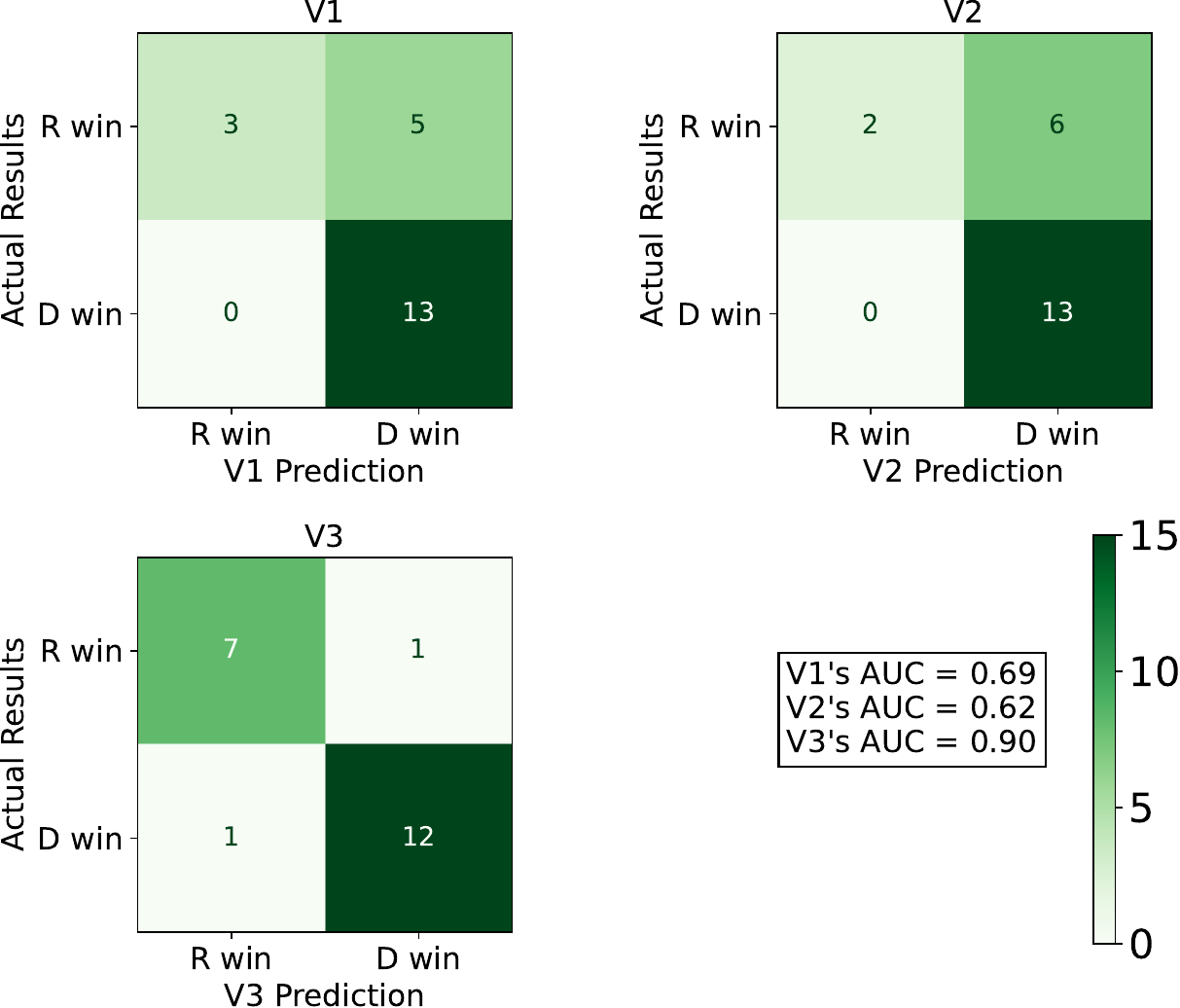}  
    \caption{
    Aggregated results of the three pipelines (V1, V2, V3) on state-level simulations. Each confusion matrix presents the number of states where predictions align with or deviate from actual outcomes. V1 (AUC = 0.69) and V2 (AUC = 0.62) show lower accuracy, while V3 (AUC = 0.90) performs best, effectively capturing Republican victories without compromising Democratic predictions. It is worth noting that, so far, we have only tested the pipelines in 21 states. If the scope is expanded to include all states, the AUC of V3 is expected to improve further, while the AUC of V1 and V2 are expected to decline.
    }
    \vspace{-0.2in}
    \label{fig:validation-state}  
\end{figure}


Consistent with the ANES dataset evaluations, the V1 pipeline (Demographic-only Prompt) exhibited a skew toward the Democratic Party, even in traditionally Republican-leaning states like South Carolina (SC), Alabama (AL), and Ohio (OH), with predictions diverging significantly from actual results. This illustrates the limitations of using demographic data alone without time-sensitive context.
The V2 pipeline (Time-dependent Prompt) introduced election-year-specific information, which partially reduced the skew in the state-level simulations. However, the model still struggled to eliminate prediction biases, particularly in polarized states. Interestingly, this differed from the ANES evaluations, where including time-dependent information amplified the bias.
The V3 pipeline (Multi-step Reasoning) demonstrated the most accurate performance, effectively mitigating skewness across deep red and blue states. In these polarized states, the predictions closely mirrored the actual voting outcomes, reflecting the model’s improved ability to incorporate ideological alignment through multi-step reasoning.

For swing and tipping-point states, the V3 pipeline achieved robust results, correctly predicting the outcomes in 9 out of 11 states. Minor deviations were observed in North Carolina (NC) and Arizona (AZ), where the predictions were slightly misaligned with the real results. Nonetheless, the V3 pipeline provided balanced predictions that accurately captured the competitive dynamics typical of swing states, further validating its effectiveness.

In summary, the comparative performance of the three pipelines across different state categories is shown in Figure~\ref{fig:example-figure}. The V3 pipeline consistently outperformed the other two, delivering more stable and accurate predictions. Aggregate results for all pipelines on all 21 chosen states is shown in the below figure ~\ref{fig:validation-state}. And all state-level simulation results can be found in the appendix.

%% file: 05conclusion.tex
\section{Conclusion and Future Directions}

In this work, we present a novel framework for election prediction using large language models (LLMs) with a focus on multi-step reasoning. 
By leveraging both synthetic personas and real-world datasets, we demonstrated the potential of LLMs to capture individual voting behaviors and state-level election outcomes. 
Our iterative design highlights the importance of integrating temporal information and complex reasoning for accurate predictions. 
The 2020 and 2024 simulations reveal both the strengths and limitations of using LLMs in dynamic political environments, emphasizing the model’s ability to generalize on unseen data while showing the challenges associated with static demographic assumptions and limited real-time data inputs. 

Future research can extend this work by incorporating multiple LLMs to better understand their internal political tendencies, enhancing temporal modeling with public opinion data and real-time trends for improved accuracy, and developing stronger multi-step reasoning pipelines through refined Chain of Thought (CoT) designs to further enhance prediction performance and mitigate biases. This study lays the foundation for future applications of LLMs in political forecasting, offering promising directions for further development in both election prediction and the broader study of LLM behavior in social contexts.